\documentclass[lettersize,journal]{IEEEtran}
\usepackage{amsmath,amsfonts}
\usepackage{algorithmic}
\usepackage{algorithm}
\usepackage{array}
\usepackage[caption=false,font=normalsize,labelfont=sf,textfont=sf]{subfig}
\usepackage{textcomp}
\usepackage{stfloats}
\usepackage{url}
\usepackage{verbatim}
\usepackage{graphicx}
\usepackage{cite}
\usepackage{amsmath,amssymb,amsfonts}
\usepackage{algorithmic}
\usepackage{graphicx}
\usepackage{textcomp}
\usepackage{wrapfig}

\usepackage{epsfig} 
\usepackage{amsmath} 
\usepackage{multirow} 
 
\usepackage{float}    
\usepackage{xcolor}

\usepackage{tabularx}
\usepackage{booktabs} 
\usepackage{array}
\usepackage{makecell}
\newcolumntype{P}[1]{>{\centering\arraybackslash}m{#1}}
\usepackage[normalem]{ulem}
\useunder{\uline}{\ul}{}
\usepackage{xcolor} 

\usepackage{makecell} 

\usepackage{titlesec}
\titlespacing{\section}{3pt}{8pt}{5pt}
\titlespacing{\subsection}{0pt}{8pt}{3pt}
\makeatletter
\renewcommand\subsubsection{\@startsection{subsubsection}{3}{\parindent}
	{0.6ex \@plus 0.2ex \@minus .2ex}
	{-0.6em}
	{\normalfont\normalsize\itshape}}
\makeatother

\begin{document}

\title{
RadarXFormer: Robust Object Detection via Cross-Dimension Fusion of 4D Radar Spectra and Images for Autonomous Driving
}

\author{Yue Sun$^{1}$, Yeqiang Qian$^{*, 2,3}$, Zhe Wang$^{4,5}$, Tianhui Li$^{4,5}$, Chunxiang Wang$^{2,3}$ and Ming Yang$^{*, 2,3}$ 
	\thanks{
		$\dag$ This work was supported by the Guangxi Science and Technology Major Program,
		“Research and Application of Digital-Twin Based Intelligent Driving Simulation and Testing Technologies” (No. AA24206056),
		and by the National Natural Science Foundation of China (Grant 62473253).
	}
	\thanks{
		$^{1}$ Global Institute of Future Technology, Shanghai Jiao Tong University, Shanghai, 200240, China.
	}
	\thanks{
		$^{2}$ School of Automation and Intelligent Sensing, Shanghai Jiao Tong University, Shanghai, 200240, China.
	}
	\thanks{
		$^{3}$ Key Laboratory of System Control and Information Processing, Ministry of Education of China, Shanghai, 200240, China.
	}
	\thanks{
		$^{4}$ SAIC GM Wuling Automobile Company Co., Ltd., Liuzhou, 545007, China.
	}
	\thanks{
		$^{5}$ Guangxi Laboratory of New Energy Automobile, Liuzhou, 545007, China.
	}
	\thanks{
		$^{*}$ Corresponding authors: Yeqiang Qian; Ming Yang.
	}
}

%

\maketitle

\begin{abstract}
Reliable perception is essential for autonomous driving systems to operate safely under diverse real-world traffic conditions. 
However, camera- and LiDAR-based perception systems suffer from performance degradation under adverse weather and lighting conditions, 
limiting their robustness and large-scale deployment in intelligent transportation systems. 
Radar-vision fusion provides a promising alternative by combining the environmental robustness and cost efficiency of millimeter-wave (mmWave) radar with the rich semantic information captured by cameras. 
Nevertheless, conventional 3D radar measurements lack height resolution and remain highly sparse, 
while emerging 4D mmWave radar introduces elevation information but also brings challenges such as signal noise and large data volume. 
To address these issues, this paper proposes RadarXFormer, 
a 3D object detection framework that enables efficient cross-modal fusion between 4D radar spectra and RGB images. 
Instead of relying on sparse radar point clouds, 
RadarXFormer directly leverages raw radar spectra and constructs an efficient 3D representation that 
reduces data volume while preserving complete 3D spatial information. 
The “X” highlights the proposed cross-dimension (3D-2D) fusion mechanism, in which 
multi-scale 3D spherical radar feature cubes are fused with complementary 2D image feature maps. 
Experiments on the K-Radar dataset demonstrate improved detection accuracy and robustness under challenging conditions while maintaining real-time inference capability.
\end{abstract}


\section{Introduction}
\label{sec:introduction}
\IEEEPARstart{S}{afety} 
and reliability are fundamental requirements for the transition of autonomous driving systems from experimental demonstrations to large-scale deployment in intelligent transportation systems. 
In real-world traffic scenarios, perception systems must not only accurately recognize objects under normal conditions 
but also maintain stable and reliable performance under diverse and challenging conditions, 
including nighttime driving, extreme weather, and adverse lighting conditions where perception failures may directly lead to safety risks. 
Ensuring robust perception under such conditions remains one of the key challenges limiting the practical deployment of autonomous vehicles.
Current autonomous driving perception systems predominantly rely on cameras or LiDAR sensors. 
While these sensors provide dense environmental observations under favorable conditions, 
their performance can degrade significantly in challenging lighting environments such as strong sunlight, backlighting, or nighttime scenarios. 
Moreover, high-performance LiDAR sensors incur substantial hardware costs, 
which restrict large-scale deployment in cost-sensitive intelligent transportation systems. 
These limitations motivate the exploration of other sensing modalities that can enhance perception reliability without significantly increasing system costs.

Millimeter-wave (mmWave) radar has recently attracted increasing attention 
due to its strong robustness under adverse environmental conditions such as rain, snow, fog, and dust. 
In addition to geometric sensing, radar measurements provide Doppler velocity information, 
which is particularly valuable for distinguishing different traffic participants \cite{modern}. 
However, radar signals are inherently sparse and noisy, 
making radar-only perception insufficient for accurate object detection. 
Fortunately, cameras, as another low-cost sensing modality, provide rich semantic information but lack reliable depth and motion sensing under degraded conditions. 
Therefore, radar-camera fusion has emerged as a promising approach to achieve both robustness and cost efficiency in autonomous driving perception systems.
Early studies explored object detection by fusing conventional 3D mmWave radar with cameras, 
where radar information is typically represented as sparse point clouds \cite{centerfusion,craft,bevfusion}. 
However, these approaches often achieve limited performance due to the inherent sparsity and lack of height resolution in 3D radar measurements. 
Furthermore, additional information loss is introduced during point cloud extraction processes such as Constant False Alarm Rate (CFAR) filtering \cite{cfar}, 
which removes weak yet informative radar signals.

With the emergence of 4D mmWave radar, which provides elevation resolution, 
recent studies have begun investigating radar-camera fusion for object detection. 
Nevertheless, most existing methods still rely on radar point clouds, 
whose sparsity continues to constrain perception performance. 
This limitation is partly due to previous public datasets such as nuScenes \cite{nuscenes} and VoD \cite{vod}, 
which only provide processed radar point cloud data. 
Since the release of the K-Radar dataset \cite{kradar},
researchers have been able to access raw radar spectrum data prior to point cloud extraction. 
Experimental results in \cite{kradar} demonstrate that directly leveraging spectral data significantly improves detection accuracy compared with point-cloud-based approaches.
Inspired by this, subsequent works have attempted to fuse radar spectra with images to further enhance perception performance. 
For instance, EchoFusion \cite{echofusion} fuses 2D radar spectrum representations (Range-Azimuth, RA maps) with images, 
while DPFT \cite{dpft} integrates both RA and EA (Elevation-Azimuth) spectral representations from 4D radar with images, 
further improving detection accuracy.
However, these methods typically compress high-dimensional 4D radar spectra into 2D representations before fusion, 
which limits the exploitation of 3D spatial information from radar data \cite{echofusion} and 
introduces additional computational overhead due to multiple attention operations \cite{dpft}.
As a result, effective and efficient fusion between high-dimensional radar spectra and images remains an open challenge for robust autonomous driving perception.

In particular, radar-camera fusion faces several fundamental challenges. 
First, multimodal sensing introduces discrepancies in resolution, data format, sensing characteristics, and spatial alignment. 
Second, raw 4D radar spectra contain extremely large data volumes, 
for instance, each frame of the 4D radar spectrum in the K-Radar dataset occupies approximately 300 MB, 
making direct utilization computationally impractical. 
Therefore, data compression is necessary before feeding radar spectra into the model, 
yet improper compression may discard informative signals, especially those that are inherently sparse but crucial for perception performance.
Finally, the dimensional mismatch between 3D radar feature cubes and 2D image feature maps makes cross-modal fusion design particularly challenging, 
requiring effective fusion strategies that preserve complementary information while maintaining computational efficiency.

To address these challenges, we propose RadarXFormer, 
a 3D object detection framework based on cross-dimension feature fusion of 4D mmWave radar spectra and images. 
Instead of compressing radar data into 2D representations, 
our method adopts an efficient 3D radar representation to preserve spatial structure while reducing the attention-related computational cost associated with multi-view radar fusion, as adopted in \cite{dpft}.
Specifically, 3D feature cubes are extracted from radar spectra, 
while 2D feature maps are obtained from images. 
A cross-dimension transformer with spherical 3D object queries is designed to achieve effective multimodal fusion without losing height information, which is often discarded by existing methods that integrate features in the bird’s-eye-view (BEV) space \cite{echofusion}. 
Through this design, RadarXFormer improves detection robustness and accuracy while maintaining compact model size and real-time inference capability.
The main contributions of this work are summarized as follows:
\begin{enumerate}
	\item 
	We propose RadarXFormer, a multimodal 3D object detection framework that improves perception robustness for autonomous driving 
	by effectively integrating spatial and Doppler information from 4D mmWave radar with high-resolution image semantics.
	\item 
	To address the challenges arising from high noise and large data volume in radar signals, 
	we design an efficient preprocessing method for 4D radar spectra and their efficient 3D representations. 
	We further explore a denoising scheme to reduce noise and storage demands, and assess its performance.
	\item 
	We introduce a cross-dimensional fusion mechanism that effectively integrates 3D radar features with 2D image features, enabling accurate and real-time object detection.
\end{enumerate}

\section{Related Work}

\subsection{Radar Data Representations}
MmWave radars in autonomous driving typically operate based on Frequency Modulated Continuous Wave (FMCW) technology. 
The principle is to continuously transmit chirp signals whose frequency increases linearly over time and 
to analyze the received echoes in order to estimate the range, angle, and Doppler velocity of objects \cite{texas}. 
Traditional mmWave radars equipped with horizontally arranged antenna arrays can 
measure range, Doppler, and azimuth information but lack elevation resolution, and are therefore often referred to as 3D mmWave radars.
In contrast, 4D mmWave radars have recently emerged, incorporating both horizontal and vertical antenna arrays to enable elevation measurement, 
thereby achieving full 3D spatial perception \cite{survey}.

A typical signal processing pipeline of 4D millimeter wave radar involves multiple stages, 
where radar data can be represented as raw ADC data, radar spectra, and radar point clouds. 
Specifically, each transmitter (TX) emits a sequence of FMCW chirps, 
while the receivers (RX) capture the reflected signals, mix them with the transmitted signals, and apply low-pass filtering to produce intermediate frequency (IF) signals. 
These signals are then digitized by an analog-to-digital converter (ADC) to generate the raw ADC data.

In the time domain, range, Doppler, and angular information are entangled and difficult to separate. 
After performing FFT along each dimension, these attributes can be mapped into the frequency domain and effectively decoupled \cite{texas}.
Specifically, a range FFT applied to fast-time samples within each chirp extracts range information; 
a Doppler FFT performed on slow-time samples across multiple chirps yields Doppler velocity; 
and angular FFTs or beamforming operations on the antenna array data estimates azimuth and elevation angles \cite{survey}. 
The resulting multi-dimensional data are collectively referred to as radar spectra, 
which can be represented in various forms, such as 
Azimuth-Doppler (AD) maps, Range-Doppler (RD) maps, Range-Azimuth (RA) maps, Range-Azimuth-Doppler (RAD) cubes, 
or Doppler-Range-Elevation-Azimuth (DREA) tesseracts, depending on the specific FFT dimensions applied.

In these spectra, real targets typically appear as localized intensity peaks. 
To distinguish them from noise and clutter, methods like CFAR detection \cite{cfar} 
are commonly employed to adaptively select significant peaks based on signal strength. 
These detected peaks are then mapped into 3D coordinates, ultimately forming the radar point cloud.
Such an extraction process inevitably results in information loss, further degrading the already sparse radar signals. 

Based on the above discussion, unlike most existing methods that rely on radar point clouds as input, 
our approach leverages radar spectrum prior to CFAR processing to preserve richer and more informative cues for perception.

\subsection{Radar Feature Extraction}
The methods for radar feature extraction vary depending on the data representation of mmWave radar. 
Here, we summarize existing radar feature extraction approaches according to different data formats.

Currently, most radar feature extraction methods are based on radar point clouds. 
This is primarily because most publicly available radar datasets, such as the 3D radar based nuScenes \cite{nuscenes} and 
the 4D radar based VoD \cite{vod} and Astyx \cite{astyx}, only provide radar point cloud data. 
In addition, since LiDAR point cloud feature extraction algorithms are already well-developed, 
many radar based approaches adopt similar processing pipelines. 
Typically, radar point clouds are first voxelized or pillarized \cite{pointpillars}, followed by convolutional neural networks for feature extraction. 
Representative examples include \cite{centerfusion,bevfusion} for 3D radar, 
and \cite{rcfusion,lxl,smurf,smiformer,l4dr,interfusion} for 4D radar. 
Other methods, such as \cite{robust,ratrack}, employ PointNet++ \cite{pointnet++}, 
while approaches like \cite{lrf} use sparse convolutional networks to process radar point clouds.
However, mmWave radar point clouds are significantly sparser than LiDAR point clouds, 
and small objects may even lack detectable points. 
Moreover, radar point clouds exhibit fundamentally different physical and statistical characteristics from LiDAR point clouds, 
which limits the effectiveness of conventional point cloud based feature extraction methods when directly applied to radar data.

Although the advent of 4D mmWave radar has improved signal resolution to some extent, it does not fully resolve the sparsity problem. 
Consequently, researchers have increasingly shifted their focus toward radar spectra to exploit richer and more informative signal features. 
Meanwhile, public datasets such as RADIal \cite{radial} and K-Radar \cite{kradar} provide pre-CFAR 4D radar data, 
where RADIal provides ADC data and K-Radar provides radar spectra, facilitating research on raw radar based perception.
Algorithms based on 2D radar spectra typically utilize RD or RA maps. 
RD based methods, such as \cite{od-rt-cnn,TFFTRadNetOD,efficient-dash}, 
extract features using convolutional networks similar to image feature extraction. 
RA based approaches often employ polar-coordinate attention mechanisms to generate BEV features, 
which are then either transformed into Cartesian coordinates via bilinear interpolation \cite{radatron} or 
directly processed within polar-coordinate detection frameworks \cite{echofusion,depth-aware}.
However, features extracted from these 2D radar spectra lack complete 3D spatial information, 
which limits their effectiveness for accurate 3D perception.
In contrast, algorithms utilizing 3D radar spectrum are relatively few. 
\cite{transradar} takes the RAD cube as input but first projects it onto AD, RD, and RA planes, 
and then encodes the features using adaptive directional attention blocks.
\cite{dpft} compresses the 4D radar spectrum data onto the RA and EA planes, 
extracts features using 2D convolutions, and performs multi-stage attention based fusion with image features.
Such methods essentially extract features from multiple 2D radar spectra viewed from different perspectives and fuse them through repeated attention operations, 
which inevitably increases computational complexity.
Other approaches, such as \cite{enhancedkradar,kradar,rtnh2}, 
first extract 3D features from 3D radar spectra represented in Cartesian coordinates using simple sparse 3D convolutional blocks.
Among them, \cite{kradar,enhancedkradar} merge the vertical and channel dimensions and apply a 2D deconvolutional network to generate BEV features, 
whereas \cite{rtnh2} introduces a dedicated vertical encoding network for BEV transformation.
However, these methods primarily exploit the 3D spatial information of radar measurements without incorporating Doppler cues.  
Moreover, converting the data into Cartesian coordinates may discard motion-related information and introduce additional errors during coordinate transformation.

In addition, radar ADC signals, as the rawest form of radar data, preserve complete reflected signal information. 
However, their extremely large data volume leads to high computational and storage costs \cite{lack_adc_irregular}, 
so only few methods attempt to extract features directly from ADC data. 
Some studies, such as \cite{TFFTRadNetOD} and \cite{adcnet}, simulate Fourier transforms using learnable fully connected layers to convert ADC data into RD maps or RAD cubes. 
However, these learned FFT operations can introduce additional bias and degrade spectral accuracy.

\begin{figure*}[!t]
	\centerline{
		\includegraphics[width=0.95\textwidth]{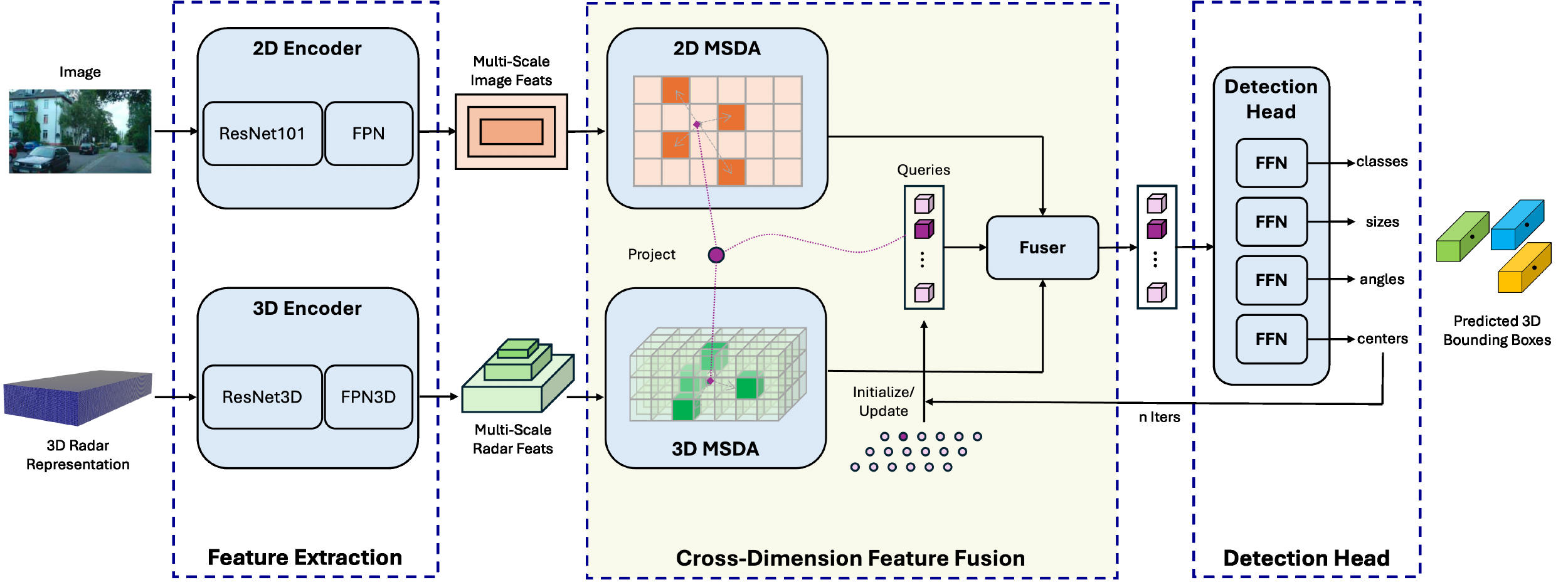}
	}
	\caption{The overall architecture of RadarXFormer, 
		where multi-scale 3D radar and 2D image features are extracted by the encoders in the Feature Extraction module, 
		fused with 3D object queries through multi-scale deformable attention in the Cross-Dimension Feature Fusion module, 
		and further refined via iterative query and detection processes for accurate 3D object estimation.}
	\label{framework}
\end{figure*}
\vspace{-5pt}

\subsection{Radar-Camera Fusion}
According to the stage of fusion, 
radar and camera fusion can be categorized into data-level, feature-level, and object-level fusion, 
consistent with other multimodal fusion approaches.

Data-level fusion integrates raw or pre-processed data from radar and camera sensors at an early stage of the perception pipeline. 
For example, \cite{deep-rc} concatenates camera images and radar point clouds as inputs to a VGG network \cite{vgg} for joint feature extraction. 
\cite{radsegnet} combines camera semantic maps, radar point clouds, and BEV grid maps to assist in identifying radar points corresponding to objects of interest. 
\cite{rrpn} generates object proposals from radar point clouds to constrain the detection scope in images, 
while \cite{rc-pixel} associates radar points with neighboring image pixels to enhance and densify radar representations.
Data-level fusion is highly sensitive to temporal and spatial misalignment, 
making precise extrinsic calibration critical for maintaining reliable correspondence between radar and image data. 
Due to these challenges, data-level fusion is not the most effective or robust fusion strategy for practical deployment.

Feature-level fusion combines features extracted independently from radar and camera data at an intermediate stage of the network. 
In \cite{bevfusion,lxl,distant}, radar and camera features are fused through concatenation or addition. 
\cite{centerfusion} employs CenterNet \cite{centernet} to localize object centers in images and then 
applies a frustum based association strategy to align radar detections with corresponding image objects, 
generating radar feature maps that augment the image features. 
Other approaches, such as \cite{echofusion,rcfusion,spatial-attn},  
utilize attention mechanisms to integrate radar and camera features in the BEV feature space, 
while \cite{dpft} projects queries into 2D radar and image feature maps for cross-modal fusion.
Feature-level fusion offers distinct advantages, 
as it enables the use of modality-specific feature extractors while enabling neural networks to learn complementary representations across different sensing modalities.
For this reason, feature-level fusion is adopted in our approach.

Object-level fusion integrates independently detected objects from radar and camera at the final stage to produce the overall perception results. 
For instance, \cite{od-navi} projects radar detections onto the image plane and aligns them with camera-detected objects, 
while \cite{rc-rep} learns semantic representations from both modalities and enhances association accuracy by 
minimizing the Euclidean distance between radar point clouds and image bounding boxes. 
Although widely adopted in traditional radar and camera fusion systems, 
object-level fusion heavily depends on the accuracy of individual detectors. 
Moreover, rich intermediate features are discarded, which limits the amount of information available for decision-making.


\section{Methodology}\label{method}

\subsection{System Overview}
The proposed RadarXFormer framework consists of four main modules: 
data preprocessing, multi-scale feature extraction, cross-dimension feature fusion, and a 3D object detection head.
The raw 4D radar spectrum is first transformed into an efficient 3D representation through data preprocessing.
Subsequently, a 3D encoder and a 2D encoder are employed to extract multi-scale radar feature cubes and image feature maps, respectively.
A set of 3D object queries is initialized within the 3D regions of interest and updated with cross-dimension features from both modalities through multi-scale deformable attention.
These queries are iteratively refined based on the predicted object centers to achieve accurate 3D object detection.
An overview of the proposed RadarXFormer system is illustrated in Fig.~\ref{framework}, 
and the details of each module are explained in the following.

\subsection{Data Preprocessing}\label{data_preprocess}

\begin{figure}[h]
	\centerline{
		\includegraphics[width=0.49\textwidth]{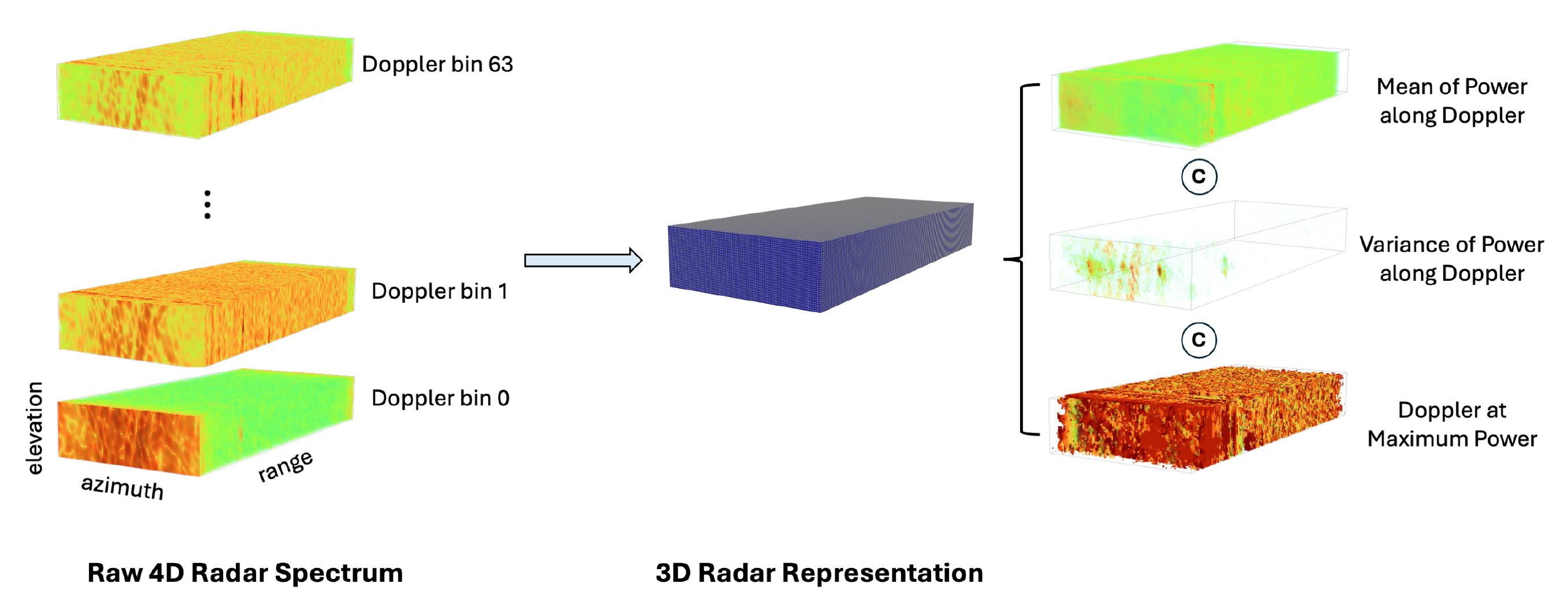}
	}
	\caption{Visualization of an example from the K-Radar dataset showing the raw 4D radar spectrum and the proposed 3D radar representation.
		The symbol \textcircled{c} indicates channel-wise concatenation.}
	\label{rt_data}
\end{figure}

We adopt the 4D radar spectrum as the radar input, which is stored as a 4D tensor (as shown on the left side of Fig.~\ref{rt_data}). 
It is extremely large (approximately 300 MB per frame in the K-Radar dataset), making direct usage impractical.
To overcome these issues, we design a radar data preprocessing pipeline that 
compresses 4D radar spectrum while preserving essential information.
The detailed process is as follows.

\begin{figure}[ht]
	\centerline{
		\includegraphics[width=0.45\textwidth]{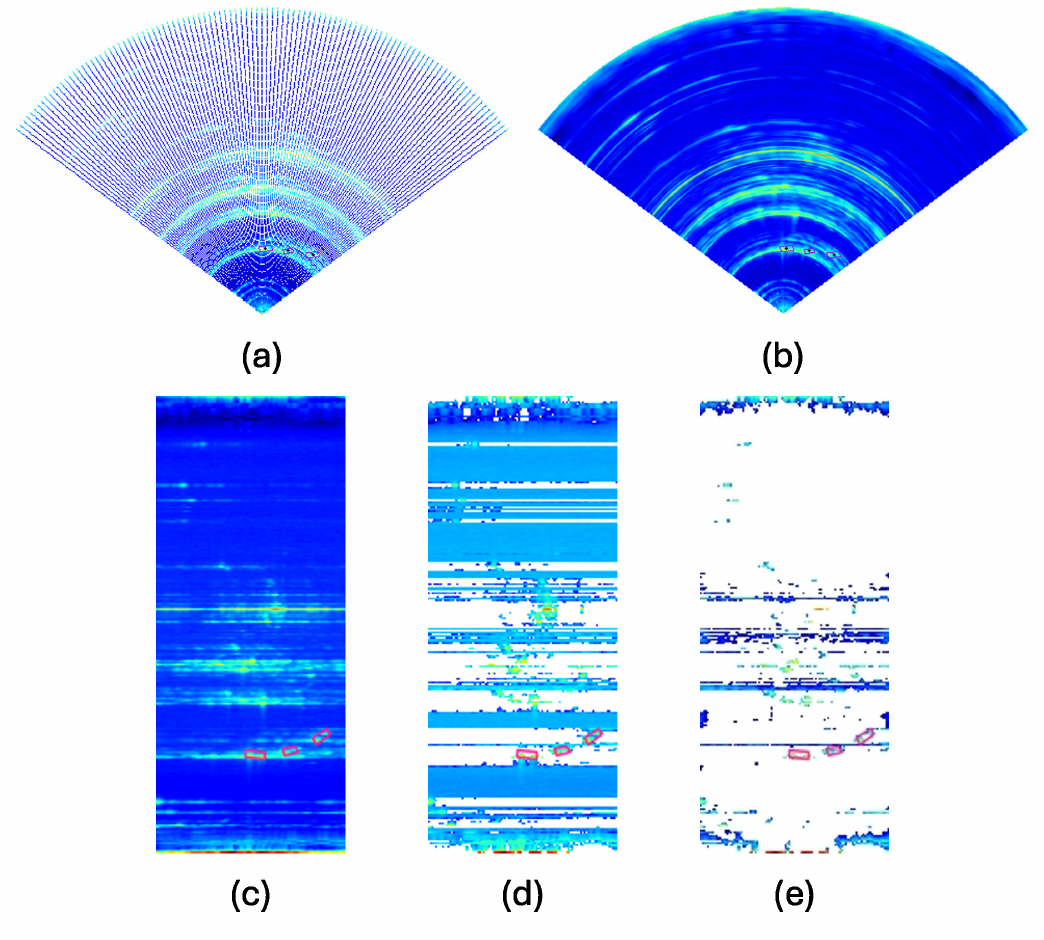}
	}
	\caption{Visualization of the same radar spectrum in BEV under different coordinates and processing stages. 
		(a) Radar spectrum in Cartesian coordinates without interpolation; 
		(b) Cartesian spectrum with interpolation; 
		(c) raw RA map in spherical coordinates; 
		(d) RA map after coarse filtering; 
		(e) RA map after coarse filtering and CFAR. 
		Pink boxes indicate ground-truth objects.}
	\label{bev}
\end{figure}

Let the original radar spectrum be represented in the spherical coordinate system as $\mathcal{R} \in \mathbb{R}^{D \times R \times E \times A}$, 
where $D$, $R$, $E$ and $A$ denote the number of bins along the Doppler, range, elevation, and azimuth dimensions, respectively. 
Unlike prior works \cite{echofusion,vd-rad} that convert radar signals into the Cartesian coordinate system before feature encoding, 
we preserve the radar data in the spherical coordinate system (its BEV shown in Fig.~\ref{bev}(c)), 
which avoids long-range sparsity and interpolation errors caused by Cartesian transformation, as shown in Fig.~\ref{bev}(a) and (b).
To retain the full 3D spatial representation while reducing dimensionality, 
the Doppler axis is statistically compressed by computing the mean and standard deviation of the power values, 
following prior studies \cite{dpft,radarocc} and experimental evaluation on a small subset of the dataset. 
In addition, the Doppler values corresponding to the maximum power peaks are recorded, 
thereby providing velocity cues for distinguishing different objects. 
The resulting features are defined as follows:
\begin{align}
	\mathcal{M} &= \frac{1}{D} \sum_{d_i=1}^{D} \mathcal{T}(d_i, :, :, :), \\[6pt]
	\mathcal{V} &= \frac{1}{D} \sum_{d_i=1}^{D} \left(\mathcal{T}(d_i, :, :, :) - \mathcal{M}\right)^{2}, \\[6pt]
	\mathcal{D} &= Doppler[\operatorname*{arg\,max}_{d_i} \mathcal{T}(d_i, :, :, :)]
\end{align}
We then concatenate these three 3D radar tensors along the channel dimension to form a new radar representation as 
\begin{equation}
	\mathcal{T} = \mathrm{Concat}(\mathcal{M}, \mathcal{V}, \mathcal{D})
	\in \mathbb{R}^{R \times E \times A \times 3}
\end{equation}

This reduces the data volume by a factor of $D/3$, where $D$ denotes the Doppler dimension (64 in K-Radar), 
while preserving the essential Doppler information. 
The visualization of the resulting 3D radar representation is shown in Fig.~\ref{rt_data}.
This preprocessing efficiently reduces the data volume of the 4D radar spectrum, 
thereby lowering storage and computational costs while preserving complete 3D spatial information. 
Unlike previous state-of-the-art 4D radar spectrum based methods \cite{echofusion} and \cite{dpft}, 
it maintains spatial integrity and enhances feature robustness.

In addition, we further explore a denoising method to suppress noise and further reduce data size, 
and evaluate its effectiveness, with detailed discussions presented in the ablation study.

\subsection{Multi-Scale Feature Extraction}
Before performing multimodal fusion, we first extract expressive multi-scale features from both the 3D radar representation and the 2D RGB image 
to ensure that the extracted features are suitable for detecting objects of varying scales. 
To this end, both the radar and image branches adopt a two-stage feature extraction pipeline consisting of a backbone network and a neck network, 
which together generate hierarchical features for subsequent fusion.

For the radar branch, a ResNet3D with depth 18 is employed as the backbone to extract multi-scale feature cubes from the 3D radar representations with three channels, 
denoted as $\mathcal{T} \in \mathbb{R}^{R \times E \times A \times 3}$. 
These feature cubes are subsequently fed into a 3D Feature Pyramid Network (FPN3D), 
which serves as the neck to unify feature dimensions and enhance the representation capability across different spatial scales.

For the image branch, given an input RGB image $\mathcal{I} \in \mathbb{R}^{H \times W \times 3}$, 
we adopt the same backbone and neck architectures as in \cite{dpft} to extract multi-scale image feature maps. 
Skip connections are also employed to directly propagate the input features to the neck.

After this step, we obtain multi-scale 3D radar feature cubes and multi-scale 2D image feature maps, 
which serve as inputs to the subsequent cross-dimension feature fusion module.

\subsection{Cross-Dimension Feature Fusion}
\begin{figure}[!t]
	\centerline{
		\includegraphics[width=0.45\textwidth]{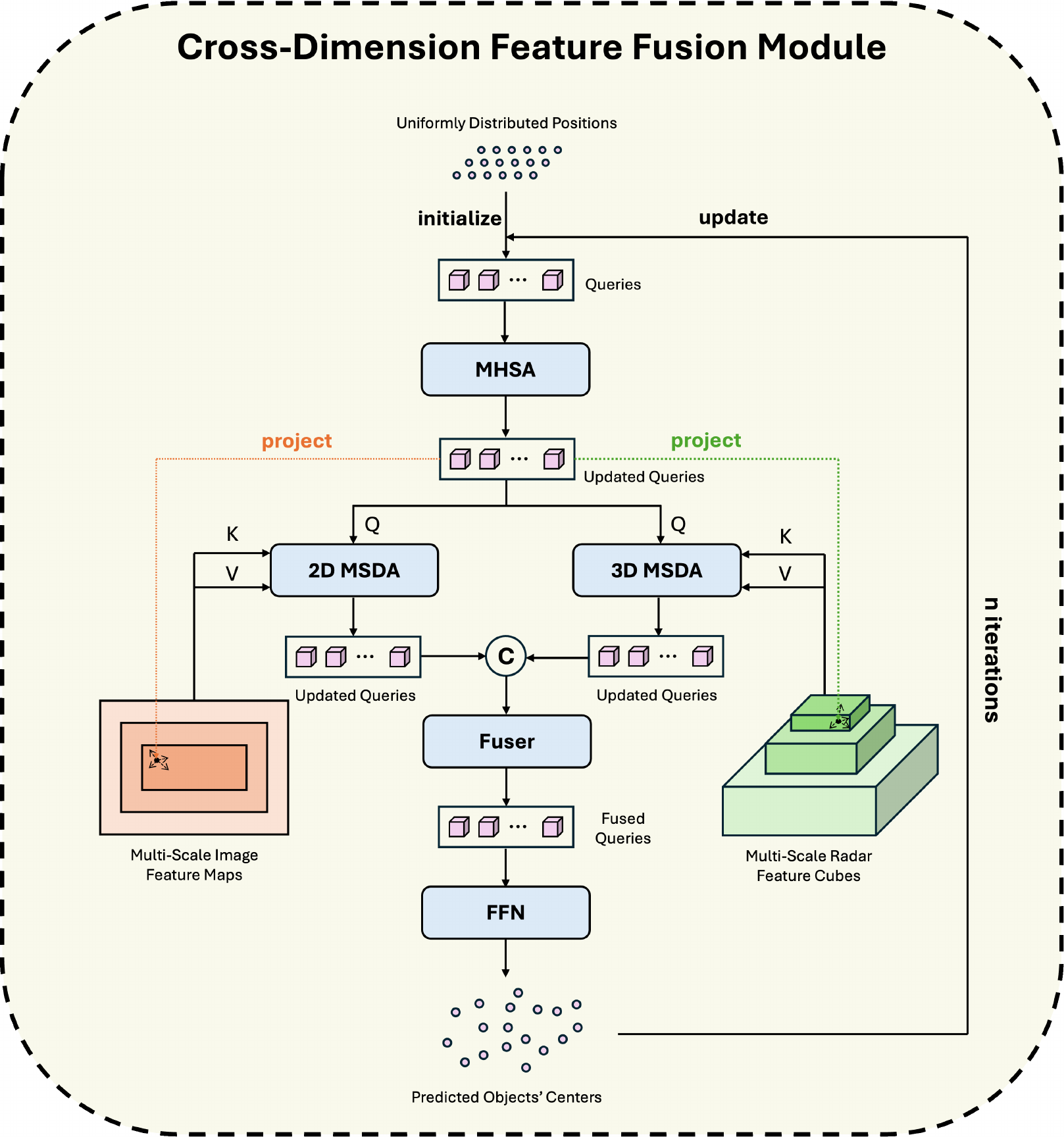}
	}
	\caption{Design details of the Cross-Dimension Feature Fusion module. 
		MHSA denotes multi-head self-attention, while MSDA denotes (multi-head) multi-scale deformable attention.}
	\label{cross-dim}
\end{figure}

The design details of the proposed cross-dimension feature fusion module are illustrated in Fig.~\ref{cross-dim}. 
This module enables direct querying of features from each input modality and retrieves object information through multi-head multi-scale deformable attention (MSDA) \cite{deformable}, 
which implicitly integrates radar and image features. 
Consequently, an explicit intermediate feature space is unnecessary and can effectively mitigate fusion errors caused by inaccurate sensor intrinsics and extrinsics.

Specifically, similar to \cite{dpft}, 
we first use a set of $N$ uniformly distributed 3D reference points within the full field of view (FoV) in the spherical coordinate system to initialize a set of queries. 
These queries are then fed into a multi-head self-attention (MHSA) \cite{attention} layer to enable information exchange among queries. 
\begin{equation}
	\mathbf{Q'} = \mathrm{MHSA} (\mathbf{Q})
\end{equation}

Next, the reference points are projected onto both the multi-scale image feature maps and the radar feature cubes based on the sensors’ intrinsics and extrinsics. 
Multi-head multi-scale deformable cross-attention is applied to learn the offsets relative to the projected locations, 
and new features are obtained at the corresponding positions via bilinear interpolation. 
The detailed mathematical derivation is as follows.

Let the feature of a query $q \in \mathbf{Q'}$ be $f_q$ and its projected point coordinates, 
normalized within each feature level, are denoted as $\hat{p}_q$.
Given a set of multi-scale feature maps $\{x^l\}_{l=1}^{L}$, 
the MSDA process can be formulated as:
\vspace{-5pt}
\begin{equation}
	\begin{split}
		\mathrm{MSDA}(f_{q}, \hat{p}_{q}, \{x^l\}_{l=1}^{L}) = 
		& \sum_{m=1}^{M} \mathbf{W}_m \Bigg(
		\sum_{l=1}^{L} \sum_{k=1}^{K}  \, \\
		& A_{mlqk} \, \mathbf{W}'_m x^l \big( \phi_l(\hat{p}_q) + \Delta p_{mlqk} \big)
		\Bigg)
	\end{split}
\end{equation}
where $\mathbf{W}_m$ and $\mathbf{W}'_m$ are the learnable weight matrices of the $m$-th attention head, 
and $\mathbf{A}_{mlqk}$ and $\Delta p_{mlqk}$ denote the attention weight and the sampling offset 
for the $k$-th sampling point of query $q$ in the $m$-th head and the $l$-th feature level, computed based on $f_q$. 
$\phi_l(\cdot)$ is used to map the normalized coordinates to the corresponding feature level.

The query updated by radar feature cubes and image feature maps are:
\begin{equation}
	{f}_R^q = \mathrm{MSDA} (f^q, \hat{p}_{R}^q, \mathbf{F_{R}})
\end{equation}
\begin{equation}
	{f}_I^q = \mathrm{MSDA} (f^q, \hat{p}_{I}^q, \mathbf{F_{I}})
\end{equation}
where $\hat{p}_{R}^q$ and $\hat{p}_{I}^q$ denote the normalized positions of query $q$ projected onto the radar and image coordinate systems, respectively, 
and $\mathbf{F}_{R}$ and $\mathbf{F}_{I}$ represent the multi-scale radar and image features obtained from the previous stage.
Note that unlike prior methods, since radar feature cubes are defined in the 3D spherical coordinate system, 
we extend the multi-scale deformable cross-attention module into a 3D spherical version, 
while the image feature maps remain processed with the standard 2D deformable cross-attention. 

In the final step, the queried features are concatenated and fed into a fuser network, 
where they are first processed by a feed-forward network (FFN) and then fused via a linear layer. 

Then, the fused query features are passed through a detection head composed of FFNs to predict the object bounding boxes, 
each including the class, center, size, and orientation. 
The predicted center coordinates are used to update the queries, 
and this process is repeated for $n$ iterations to progressively refine the 3D bounding box predictions for higher accuracy.

This design enables the queries to effectively bridge the gap between 3D and 2D feature representations across different dimensions.

\subsection{Detection Head and Loss}
We adopt a detection head architecture similar to that in \cite{dpft} to decode the fused query features 
for predicting the object categories and geometric attributes of the bounding boxes. 
The classification branch consists of a FFN that predicts object categories, 
while three regression branches, each implemented as an FFN, 
are used to independently predict the bounding box center coordinates $(x, y, z)$, size $(l, w, h)$, and orientation $(\sin \theta, \cos \theta)$.

During model training, a set-to-set loss based on Hungarian matching is employed, 
formulated as a weighted combination of the Focal loss for the classification branch and the $L_1$ loss for each regression branch. 
In our implementation, all loss weights are set to 1.
Thus, the overall loss function can be formulated as follows:
\vspace{-5pt}
\begin{equation}
	\mathcal{L} =\mathcal{L}_{\text{class}} +\mathcal{L}_{\text{center}}  +\mathcal{L}_{\text{size}}  + \mathcal{L}_{\text{angle}} 
\end{equation}
\vspace{-5pt}


\begin{table*}[t]
	\caption{3D object detection results (mAP, \%). The table is organized by sensor modalities, where C, L, and R denote Camera, Lidar, and Radar, respectively.}
	\label{tab:benchmark}
	\centering
	\renewcommand{\arraystretch}{1.1} %
	\setlength{\tabcolsep}{3pt}
	\begin{tabular}{@{}P{0.19\textwidth} | P{0.07\textwidth} | P{0.07\textwidth}P{0.07\textwidth}P{0.07\textwidth}P{0.07\textwidth}P{0.07\textwidth}P{0.07\textwidth}P{0.07\textwidth} | P{0.07\textwidth}@{}}
		\bottomrule[1.5pt]
		\multirow{2}{*}{Method} & \multirow{2}{*}{Modality} & \multicolumn{1}{c}{\multirow{2}{*}{Norm.}} & \multirow{2}{*}{Overcast} & \multicolumn{1}{c}{\multirow{2}{*}{Fog}} & \multicolumn{1}{c}{\multirow{2}{*}{Rain}} & \multicolumn{1}{c}{\multirow{2}{*}{Sleet}} & \multicolumn{1}{c}{\multirow{2}{*}{\begin{tabular}[c]{@{}l@{}}Light\\ Snow\end{tabular}}} & \multirow{2}{*}{\begin{tabular}[c]{@{}l@{}}Heavy\\ Snow\end{tabular}} & \multirow{2}{*}{\begin{tabular}[c]{@{}l@{}}Total\\ mAP\end{tabular}} \\
		& & & & & & & & & \\
		\hline
		RTNH~\cite{kradar}                                               & R            & 49.9   & 56.7     & 52.8 & 42.0 & 41.5  & 50.6       & 44.5       & 47.4  \\
		\hline
		Voxel-RCNN~\cite{voxel-rcnn}                           & L            & 81.8   & 69.6     & 48.8 & 47.1 & 46.9  & 54.8       & 37.2       & 46.4  \\
		CasA~\cite{CasA}                                                  & L            & \underline{82.2}   & 65.6     & 44.4 & 53.7 & 44.9  & 62.7       & 36.9       & 50.9  \\
		TED-S~\cite{TED}                                                 & L             & 74.3   & 68.8     & 45.7 & 53.6 & 44.8  & 63.4       & 36.7       & 51.0  \\
		\hline
		VPFNet~\cite{VPFNet}                                         & C + L     & 81.2   & \underline{76.3}     & 46.3 & 53.7 & 44.9  & 63.1       & 36.9       & 52.2  \\
		TED-M~\cite{TED}                                                & C + L     & 77.2   & 69.7     & 47.4 & \underline{54.3} & 45.2  & \underline{64.3}       & 36.8       & 52.3  \\
		MixedFusion~\cite{MixedFusion}                       & C + L     & \textbf{84.5}   & \textbf{76.6}     & 53.3 & \textbf{55.3} & \underline{49.6}  & \textbf{68.7}       & 44.9       & 55.1  \\
		\hline
		InterFusion~\cite{interfusion}                              & R + L     & 15.3   & 20.5    & 47.6 & 12.9 & 9.3  & 56.8       & 25.7       & 17.5  \\
		3D-LRF~\cite{lrf}                                                   & R + L     & 45.3   & 55.8    & 51.8 & 38.3 & 23.4  & 60.2       & 36.9       & 45.2  \\
		L4DR~\cite{l4dr}                                                   & R + L     & 53.0   & 64.1    & \textbf{73.2} & 53.8 & 46.2  & 52.4       & 37.0       & 53.5  \\
		\hline
		EchoFusion~\cite{echofusion}                            & R + C    & 51.5             & 65.4            & 55.0                      & 43.2         & 14.2                       & 53.4         & 40.2                       & 47.4  \\
		DPFT~\cite{dpft}                                                   & R + C     & 55.7           & 59.4            & 63.1           & 49.0   & \textcolor{blue}{\textbf{51.6}}  & 50.5         &  \textcolor{blue}{\textbf{50.5}} & \underline{56.1}  \\
		RadarXFormer (Ours)                                           & R + C    & \textcolor{blue}{56.8}  & \textcolor{blue}{66.3}  & \textcolor{blue}{\underline{67.3}}  & \textcolor{blue}{50.5} & 48.5   & \textcolor{blue}{58.5} & \underline{49.5}   & \textcolor{blue}{\textbf{57.6}}  \\
		\toprule[1.5pt]
	\end{tabular}
	\label{tab:all_methods_mAP}
\end{table*}

\section{Experiments}

\subsection{Data and Settings}\label{settings}
\subsubsection{Dataset Preparation}
To the best of our knowledge, K-Radar \cite{kradar} is the only publicly available dataset that provides raw 4D radar spectra processed solely by FFT along each dimension, 
and is specifically designed for autonomous driving applications in adverse weather environments.
Hence, we conduct all our experiments on this dataset. 
All experiments are conducted using label revision v1.0 and the official K-Radar splits, ensuring consistency with previous methods.

\subsubsection{Model Training and Testing}
All experiments are conducted on a single NVIDIA GeForce RTX 3090 GPU. 
We follow the training strategy described in \cite{dpft}, employing the AdamW optimizer with a constant learning rate of $1 \times 10^{-4}$ for up to 200 epochs.

\subsection{Evaluation Results}
All results are reported on the K-Radar~\cite{kradar} test set with label revision v1.0
and evaluated following the official protocol, which is based on the KITTI~\cite{kitti} benchmark. 
To provide a more comprehensive comparison across methods, 
we also compute errors of the predicted 3D bounding boxes, 
including position, dimensions, orientation, and both BEV and 3D translation errors. 
Since \cite{kradar} has demonstrated that models using radar spectra achieve significantly higher detection performance than those based on radar point clouds, 
we adopt EchoFusion~\cite{echofusion} and DPFT~\cite{dpft} as baseline radar-camera fusion methods, 
both of which are previous state-of-the-art approaches that directly operate on radar spectral inputs. 
As neither DPFT nor EchoFusion reports prediction error metrics, 
we compute these errors using our re-trained EchoFusion model and the official DPFT checkpoint available on GitHub. 
All other results are in line with the literature \cite{dpft}\cite{l4dr}\cite{MixedFusion}. 
The best result for each metric is highlighted in bold, the second best result is underlined, and the best performing radar-camera based method is marked in blue.

\subsubsection{Model Accuracy}

Table~\ref{tab:all_methods_mAP} presents the 3D object detection results under seven different weather conditions. 
As shown, the proposed RadarXFormer achieves the highest overall performance, 
attaining a total mean Average Precision (mAP) of 57.6\% at an IoU threshold of 0.3 for 3D bounding box detection,
surpassing all baseline methods across different sensor modalities. 
Remarkably, RadarXFormer outperforms state-of-the-art LiDAR based and camera-LiDAR fusion models overall, 
while also offering lower deployment cost. 
Although its performance is slightly lower under relatively favorable weather conditions, 
it consistently shows clear advantages across challenging scenarios. 
In particular, under severe conditions such as fog and heavy snow, 
where LiDAR or camera-LiDAR based methods suffer significant performance degradation, 
RadarXFormer demonstrates the distinctive robustness and advantages of radar sensing.
Compared with previous state-of-the-art radar and camera fusion models, EchoFusion and DPFT, 
which also take radar spectra as input, RadarXFormer improves the total mAP by 10.2\% and 1.5\%, respectively. 

\begin{table}[h]
	\centering
	\scriptsize
	\setlength{\tabcolsep}{1.5pt}
	\renewcommand{\arraystretch}{1.2}
	\caption{Detection performance comparison (mAP, \%) of different methods, reported for BEV and 3D mAP at multiple IoU thresholds.}
	\resizebox{\columnwidth}{!}{
		\begin{tabular}{>{\centering\arraybackslash}c|ccc|ccc}
			\bottomrule[1.2pt]
			\multirow{2}{*}{Method} & \multicolumn{3}{c|}{BEV mAP (\%)} & \multicolumn{3}{c}{3D mAP (\%)} \\ 
			& IoU=0.7 & IoU=0.5 & IoU=0.4 & IoU=0.7 & IoU=0.5 & IoU=0.3 \\ \hline
			\makecell[c]{RTNH}                            & 11.5                       & 43.2                    & \underline{58.4}   & 0.5                     & 15.6                      & 47.4 \\
			\makecell[c]{EchoFusion}                   & 25.7                     & 39.7                     & 48.9                       & 6.4                    & 28.1                     & 47.4 \\
			\makecell[c]{DPFT}                             & \underline{26.3} & \underline{48.5} & 57.5                       & \underline{8.0} & \underline{37.0}  & \underline{56.1} \\
			\makecell[c]{RadarXFormer\\ (Ours)} & \textbf{30.5}        & \textbf{55.6}       & \textbf{58.5}      & \textbf{10.3}      & \textbf{42.1}        & \textbf{57.6} \\
			\toprule[1.2pt]
	\end{tabular}}
	\label{tab:mAP_RC_methods_diff_IoU}
\end{table}

\begin{table*}[t]
	\centering
	\renewcommand{\arraystretch}{1.1}
    \caption{3D object detection errors (center, size, rotation, translation) of radar-camera methods.}
	\renewcommand{\arraystretch}{1.1} %
	\begin{tabular}{c|ccc|ccc|c|c|c|c|c}
		\bottomrule[1.5pt]
		\multirow{2}{*}{Method} & \multicolumn{3}{c|}{Center Error (m)} & \multicolumn{3}{c|}{Size Error (m)} & {Rot. Error} & {Trans. Error} & {Trans. Error} & \multirow{2}{*}{Scale} & {Orientation} \\
		& X & Y & Z & W & L & H & (rad) & BEV & 3D & & (\textdegree) \\
		\hline
		EchoFusion                  & 0.26              & 0.24               & 0.29               & 0.18               & {\ul 0.31}               & 0.21               & 0.27       & 0.40                     & 0.54              & {\ul 0.21}               & 18.75 \\
		DPFT                            & {\ul0.24}       & {\ul0.16}        & {\ul0.21}       & {\ul 0.11}        & \textbf{0.16}           & {\ul 0.15}       & {\ul0.29}        & {\ul0.32}       & {\ul0.42}       & \textbf{0.12}        & {\ul16.76} \\
		RadarXFormer (ours) & \textbf{0.23} & \textbf{0.14} & \textbf{0.19}  & \textbf{0.10} & \textbf{0.16}        & \textbf{0.14} & \textbf{0.13} & \textbf{0.29} & \textbf{0.38} & \textbf{0.12} & \textbf{7.19} \\
		\toprule[1.5pt]
	\end{tabular}
	\label{tab:error}
\end{table*}
Table~\ref{tab:mAP_RC_methods_diff_IoU} provides a comprehensive comparison between the proposed RadarXFormer and 
previous state-of-the-art radar or radar-camera fusion approaches. 
Across all IoU thresholds, including IoU = 0.7 (hard), IoU = 0.5 (moderate), and IoU = 0.3 (easy), 
RadarXFormer consistently achieves the highest BEV and 3D mAP. 
In particular, at IoU = 0.5, which is commonly used for object detection in driving scenarios, 
RadarXFormer outperforms EchoFusion and DPFT in 3D mAP by 14\% and 5.1\%, respectively,
demonstrating its superior capability in exploiting the complementary characteristics of radar and camera modalities for accurate and robust 3D perception.

\subsubsection{Model Prediction Error}
Table~\ref{tab:error} provides a quantitative comparison of 3D object detection errors among radar and camera fusion based methods. 
As shown, RadarXFormer achieves the lowest errors across all metrics, demonstrating its superior precision and stability. 
Specifically, RadarXFormer reduces the average center error in the X, Y, and Z directions to 0.23 m, 0.14 m, and 0.19 m, respectively, 
showing clear improvements over DPFT and EchoFusion. 
Similarly, the size estimation errors in width, length, and height are significantly smaller, indicating more accurate object boundary localization. 
In terms of rotation and translation, RadarXFormer also achieves the best performance and consistently outperforms all comparison methods. 
Notably, it attains an orientation error of only 7.19\textdegree, which is less than half that of EchoFusion and markedly lower than DPFT, 
validating the robustness of the proposed cross dimension fusion strategy in maintaining geometric consistency and orientation accuracy.

It is also worth noting that both RadarXFormer and DPFT achieve substantially smaller errors than EchoFusion, particularly along the Z axis. 
This is because EchoFusion uses only the RA map representation, which severely compresses spatial information in the height dimension. 
DPFT improves performance by incorporating EA maps but still projects radar features into 2D space. 
RadarXFormer further surpasses DPFT by leveraging complete 3D radar spectral information, 
preserving richer spatial cues and achieving more precise 3D perception, while reducing attention-related computation, 
as it fuses a unified 3D radar feature cube with a single attention operation instead of two in DPFT.

\subsubsection{Visualization of Model Output}
\begin{figure*}[!t]
	\begin{tabular}{@{}P{0.07\textwidth}P{0.03\textwidth}P{0.07\textwidth}P{0.07\textwidth}P{0.07\textwidth}P{0.07\textwidth}@{}}
		Day & \multicolumn{5}{c}{
			\begin{minipage}[c]{1.7\columnwidth}
				\centering
				\includegraphics[width=1.1\linewidth]{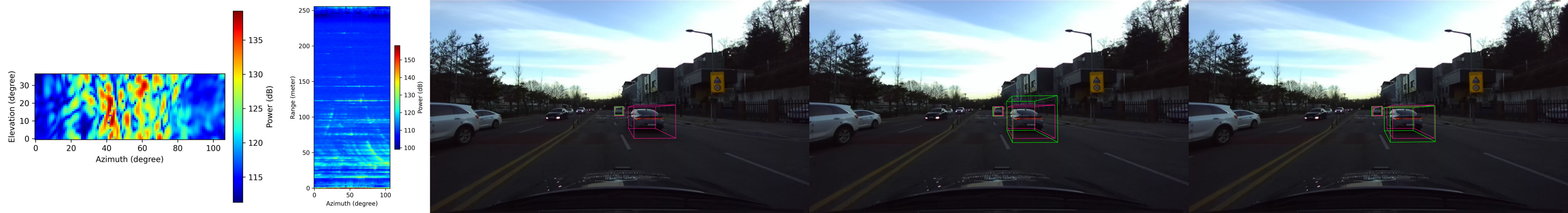}
			\end{minipage}
		}                              
		\\
		Night & \multicolumn{5}{c}{
			\begin{minipage}[c]{1.7\columnwidth}
				\centering
				\includegraphics[width=1.1\linewidth]{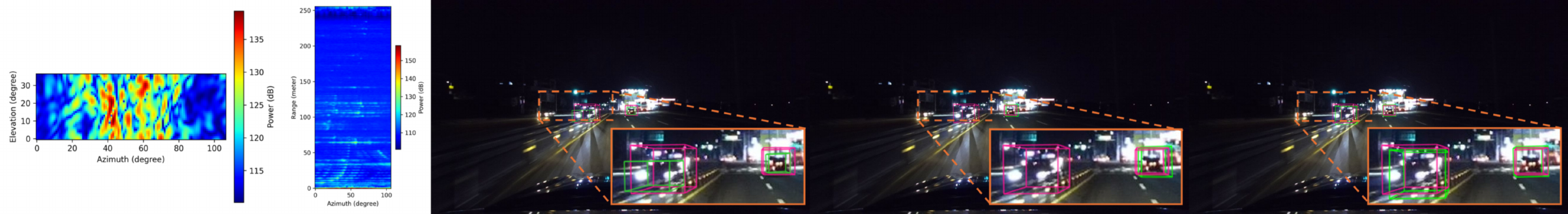}
			\end{minipage}
		}                              
		\\
		Overcast & \multicolumn{5}{c}{
			\begin{minipage}[c]{1.7\columnwidth}
				\centering
				\includegraphics[width=1.1\linewidth]{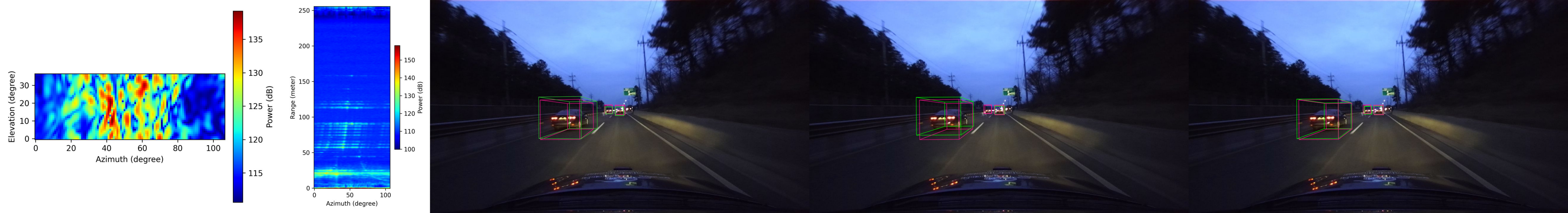}
			\end{minipage}
		}                              
		\\
		Fog & \multicolumn{5}{c}{
			\begin{minipage}[c]{1.7\columnwidth}
				\centering
				\includegraphics[width=1.1\linewidth]{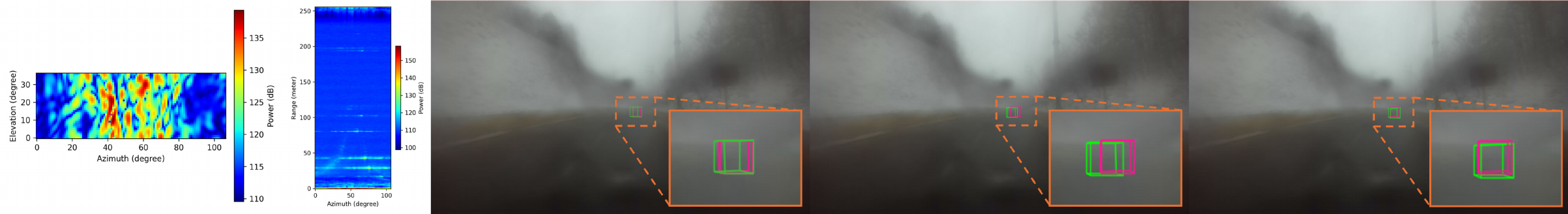}
			\end{minipage}
		}                              
		\\
		Sleet & \multicolumn{5}{c}{
			\begin{minipage}[c]{1.7\columnwidth}
				\centering
				\includegraphics[width=1.1\linewidth]{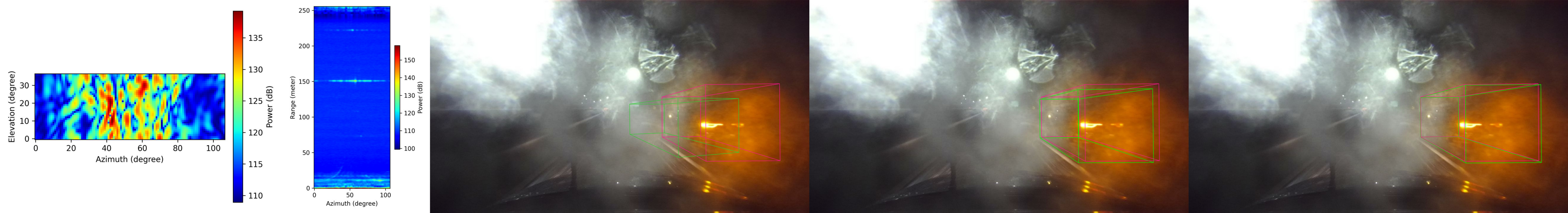}
			\end{minipage}
		}                              
		\\
		Rain  & \multicolumn{5}{c}{
			\begin{minipage}[c]{1.7\columnwidth}
				\centering
				\includegraphics[width=1.1\linewidth]{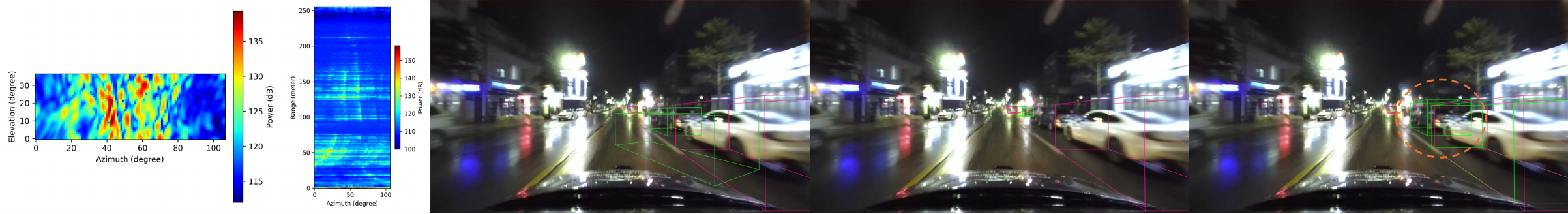}
			\end{minipage}
		}                              
		\\
		Light Snow  & \multicolumn{5}{c}{
			\begin{minipage}[c]{1.7\columnwidth}
				\centering
				\includegraphics[width=1.1\linewidth]{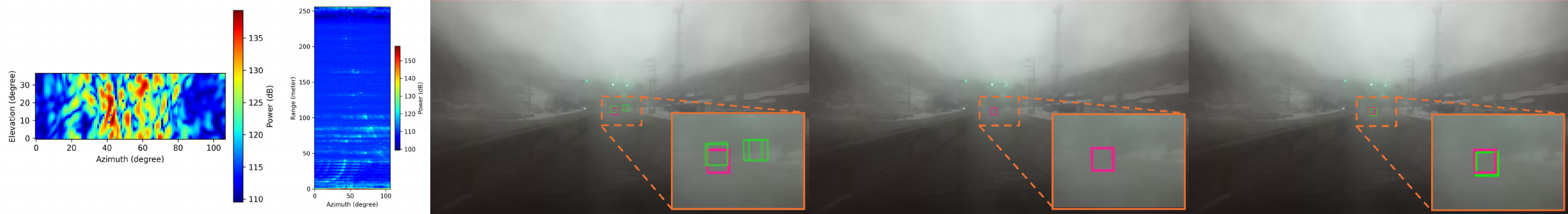}
			\end{minipage}
		}                              
		\\
		\vspace{-15000pt}Heavy Snow  & \multicolumn{5}{c}{
			\begin{minipage}[c]{1.7\columnwidth}
				\centering
				\includegraphics[width=1.1\linewidth]{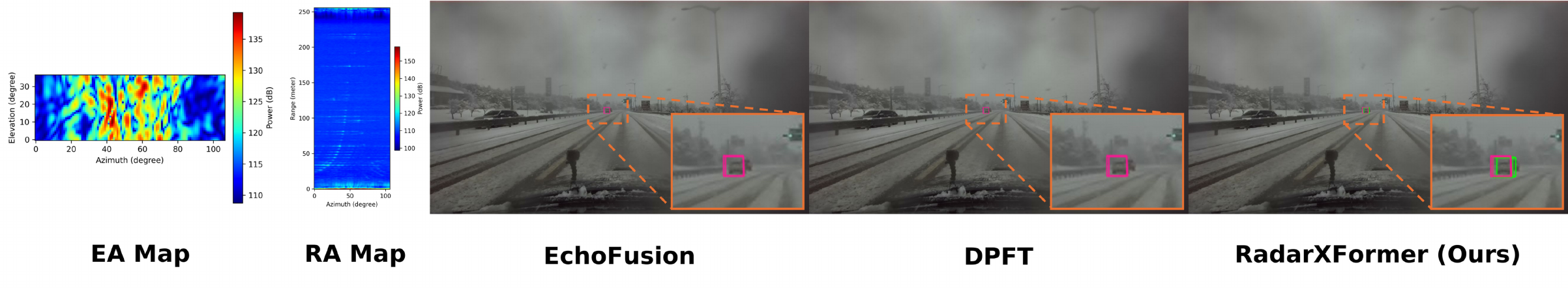}
			\end{minipage}
		}
	\end{tabular}
	\caption{Example results of radar–camera 3D object detection models under various weather conditions. 
		Ground truth and predictions are shown in pink and green. 
		EA and RA maps are displayed on the left.}
	\label{vis}
\end{figure*}

We visualize representative prediction results under various environmental scenarios, 
including \textit{day}, \textit{night}, \textit{overcast}, \textit{fog}, \textit{sleet}, \textit{rain}, \textit{light snow}, and \textit{heavy snow}, as shown in Fig.~\ref{vis}. 
The corresponding EA and RA maps are also displayed on the left for reference. 
Overall, the proposed RadarXFormer provides the most accurate object estimations across all aspects, including location, size, orientation, and detection accuracy.
In particular, for nearby objects (e.g., in \textit{sleet}, and \textit{rain} scenes), 
RadarXFormer achieves the most precise localization and oritation. 
Moreover, for distant targets, such as in the \textit{night} and \textit{snow} scenarios, other methods tend to miss detections, 
whereas RadarXFormer still successfully identifies and localizes the objects with high accuracy. 
Remarkably, RadarXFormer can even detect real objects that are not labeled in the ground truth, as observed in the \textit{rain} case, 
demonstrating its strong robustness and generalization capability under diverse and adverse weather conditions.

\subsection{Ablation Study}

\subsubsection{Effect of Noise Filtering}
We also explore a noise filtering strategy to coarsely suppress noise and clutter. 
To prevent the inadvertent removal of object reflections at long ranges due to natural signal attenuation, a range-wise filtering approach is adopted.
Specifically, for each range bin, we filter out signals whose power is lower than $\mathcal{T}_{\max} \times \alpha$ in each azimuth-elevation slice. 
In our experiments, $\alpha$ is set to 15\%. A visualization example is shown in Fig.~\ref{bev} (d).
We also show the results after applying CFAR in Fig.~\ref{bev} (e). 
Although it removes more reflections from non-object regions, 
some reflections within ground-truth boxes are mistakenly filtered out, 
causing feature loss and thereby reducing detection accuracy. 
In contrast, our method largely retains complete reflections in object regions while only leaving uniform, non-peak signals elsewhere, which barely affects object detection.

Since a large number of positions in the radar spectrum are removed after filtering, we can adopt the following sparse format to store only the remaining valid points:
\begin{equation}
	\mathcal{S} = \left\{\mathcal{P}_{r_i, e_i, a_i}, \mathcal{X}_{r_i, e_i, a_i}\big) \mid i = 1, 2, \dots, N  \right\}
\end{equation}
\begin{equation}
	\mathcal{P}_{r_i, e_i, a_i} = \big(r_i, e_i, a_i\big).
\end{equation}
\begin{equation}
	\mathcal{X}_{r_i, e_i, a_i} = \big(\mathcal{M}_{r_i, e_i, a_i}, \mathcal{V}_{r_i, e_i, a_i}, \mathcal{D}_{r_i, e_i, a_i}\big).
\end{equation}
where $r_i$, $e_i$, $a_i$ denote the range, elevation, and azimuth bin of the $i$-th point.

Note that it can also be reconstructed into the dense format:
\begin{equation}
	\mathcal{T}_{r,e,a} = 
	\begin{cases}
		\mathcal{X}_{r,e,a}, & \text{if } (r,e,a) \in \mathcal{P}, \\
		0, & \text{otherwise.}
	\end{cases}
\end{equation}

Table~\ref{tab:mAP_diff_modality} presents the detection performance of RadarXFormer under different sensor modality settings. 
As shown in the last two rows, the detection performance of the fusion model using filtered radar input is slightly lower than that using unfiltered radar input. 
However, the difference is minor, with the BEV mAP and 3D mAP decreasing by only 0.4\% and 0.6\%, respectively, at an IoU threshold of 0.3.
Both versions still outperform all other methods listed in Table~\ref{tab:all_methods_mAP} on the K-Radar dataset. 
This slight degradation may result from the inevitable loss of a few object reflection points during the filtering process.
In contrast, when using radar as a single modality (the first two rows of Table~\ref{tab:mAP_diff_modality}), 
the model with noise filtering performs better than its unfiltered counterpart. 
This can be attributed to the fact that, without complementary visual input, radar signals are more vulnerable to environmental noise. 
In such cases, the performance degradation caused by noise is more severe than the minor information loss introduced by filtering, 
making noise filtering particularly beneficial for radar-only detection.

\subsubsection{Effect of Feature Fusion}
It can be observed from Table~\ref{tab:mAP_diff_modality} that the proposed RadarXFormer achieves the highest detection performance when fusing radar and camera data. 
In particular, the fusion of unfiltered radar and camera inputs yields the best overall accuracy. 
At an IoU threshold of 0.3, the 3D mAP of the fusion model surpasses the radar-only model by 7.6\% and exceeds the camera-only model by 46.8\%, 
confirming the effectiveness of the proposed cross-dimension feature fusion in exploiting the complementary strengths of both modalities. 
Even when the radar input is filtered, the fusion model still maintains competitive performance, demonstrating strong robustness to radar signal sparsity and noise.

In the single-modality setting, both radar-only models substantially outperform the camera-only model across all IoU thresholds. 
This performance gap arises mainly because the K-Radar dataset contains numerous adverse weather scenes, 
where visual perception degrades sharply due to low visibility and lighting variation, whereas radar sensing remains stable and reliable. 

These results highlight that radar serves as a more robust sensing modality under challenging environmental conditions, 
and that its integration with vision through effective fusion brings further improvements in detection accuracy and consistency.

\begin{table}[h]
	\centering
	\scriptsize
	\setlength{\tabcolsep}{1.5pt}
	\renewcommand{\arraystretch}{1.1}
	\caption{Detection performance (mAP, \%) of RadarXFormer with different sensor modalities. 
		RF denotes radar data after noise filtering.}
	\resizebox{\columnwidth}{!}{
		\begin{tabular}{>{\centering\arraybackslash}m{0.18\columnwidth}|ccc|ccc}
			\bottomrule[1.2pt]
			\multirow{2}{*}{\centering Modality} & \multicolumn{3}{c|}{BEV mAP (\%)} & \multicolumn{3}{c}{3D mAP (\%)} \\ 
			& IoU=0.7 & IoU=0.5 & IoU=0.3 & IoU=0.7 & IoU=0.5 & IoU=0.3 \\ \hline
			\makecell[c]{R}   & 23.8      & 47.7        & 51.4       & 6.7          & 33.4      & 50.0  \\
			\makecell[c]{RF} & 24.8      & 49.2       & 52.8       & 7.0          & 28.6      & 51.1  \\
			\makecell[c]{C}   & 3.0        & 9.0          & 11.2        & 2.1          & 8.1         & 10.8  \\
			\makecell[c]{R + C} & \textbf{30.5} & \textbf{55.6} & \textbf{58.5} & \textbf{10.3} & \textbf{42.1 }& \textbf{57.6} \\
			\makecell[c]{RF + C} & \underline{29.5} & \underline{55.2} & \underline{58.1} & \underline{9.6} & \underline{41.0} & \underline{57.0} \\
			\toprule[1.2pt]
	\end{tabular}}
	\label{tab:mAP_diff_modality}
\end{table}

\subsubsection{Effect of Refinement Iterations}
Table~\ref{tab:mAP_diff_iter} presents a comparison of the detection performance of RadarXFormer using different numbers of refinement iterations. 
It can be observed that increasing the number of refinement iterations consistently improves detection performance up to four iterations, 
where RadarXFormer achieves the best BEV and 3D mAP across all IoU thresholds. 
Further increasing the number of iterations leads to performance degradation, indicating diminishing returns.
Therefore, four refinement iterations are adopted in our method.

\begin{table}[h]
	\centering
	\scriptsize
	\setlength{\tabcolsep}{1.5pt}
	\renewcommand{\arraystretch}{1.1}
	\caption{Detection performance (mAP, \%) of RadarXFormer with different refinement iterations. 
		* Our method uses 4 iterations.}
	\resizebox{\columnwidth}{!}{
		\begin{tabular}{>{\centering\arraybackslash}m{0.18\columnwidth}|ccc|ccc}
			\bottomrule[1.2pt]
			\multirow{2}{*}{\centering Modality} & \multicolumn{3}{c|}{BEV mAP (\%)} & \multicolumn{3}{c}{3D mAP (\%)} \\ 
			& IoU=0.7 & IoU=0.5 & IoU=0.3 & IoU=0.7 & IoU=0.5 & IoU=0.3 \\ \hline
			\makecell[c]{1} & 16.7  & 39.6  & 52.1  & 4.0  & 26.6  & 49.9  \\
			\makecell[c]{2} & 25.9  & 50.4  & 54.2  & 8.0  & 36.9  & 52.8\\
			\makecell[c]{3}  & {\ul28.8}         & 53.0       & 56.1                & {\ul9.5}        & {\ul39.7}         & 54.9  \\
			\makecell[c]{4*} & \textbf{30.5} & \textbf{55.6} & \textbf{58.5} & \textbf{10.3} & \textbf{42.1}    & \textbf{57.6} \\
			\makecell[c]{5}  & 28.7                & {\ul53.5}        & {\ul57.0}       & {\ul9.5}        & 39.4                  & {\ul55.9} \\ 
			\toprule[1.2pt]
	\end{tabular}}
	\label{tab:mAP_diff_iter}
\end{table}

\subsection{Discussion and Limitation}
\begin{figure}[!h]
	\centering
	\begin{minipage}[b]{0.5\columnwidth}
		\centering
		\includegraphics[width=\columnwidth]{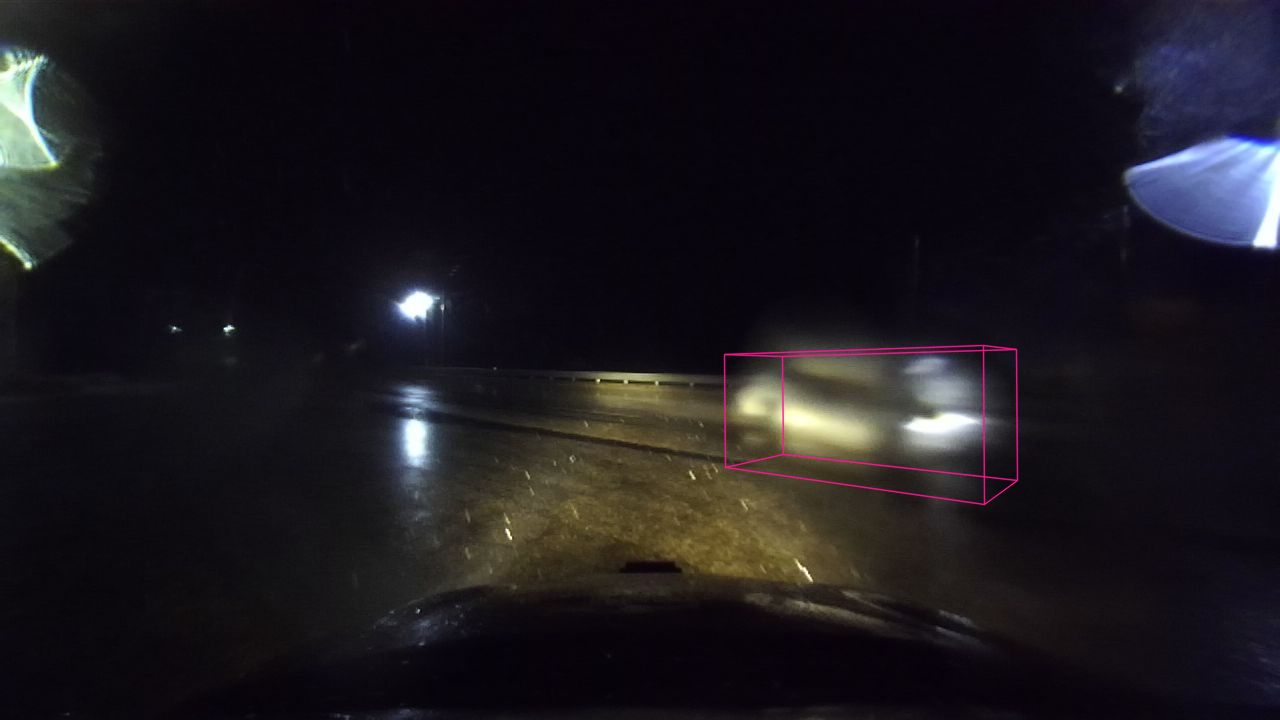}
	\end{minipage}%
	\hfill
	\begin{minipage}[b]{0.5\columnwidth}
		\centering
		\includegraphics[width=\columnwidth]{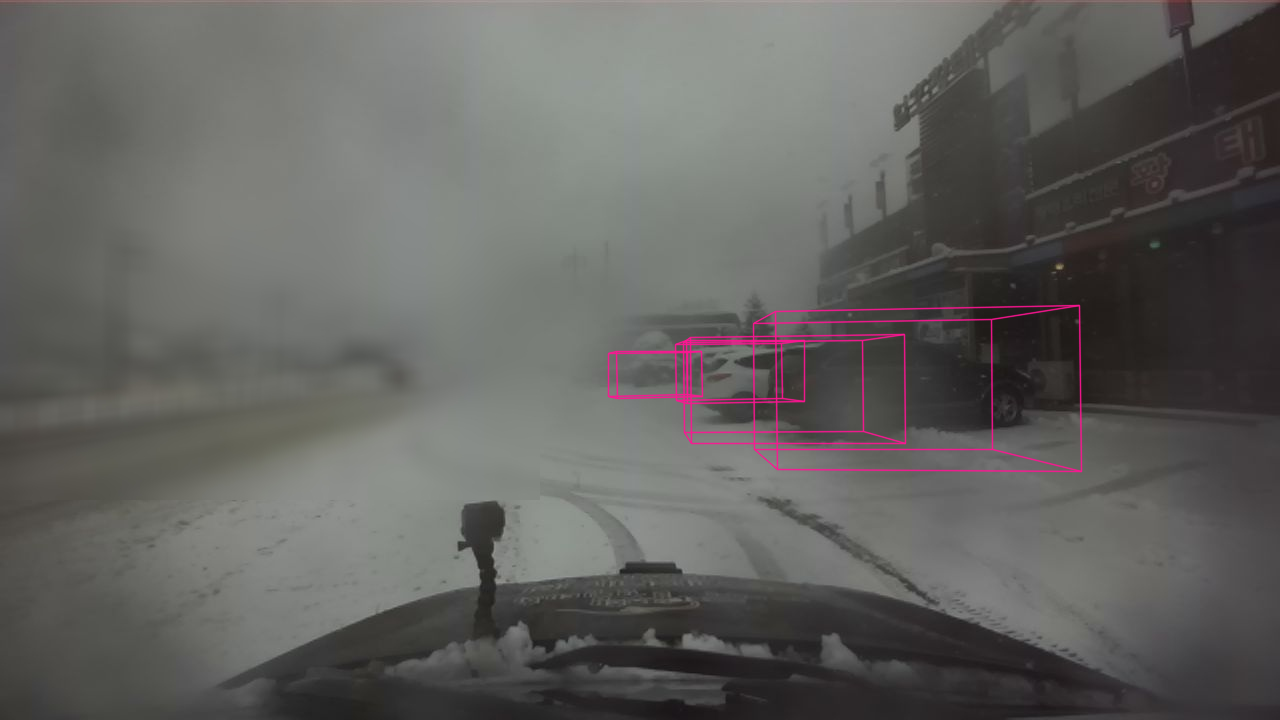}
	\end{minipage}
	\caption{Two failure cases of our model illustrating the missed detections.}
	\label{vis_failure}
\end{figure}

Fig.~\ref{vis_failure} illustrates two failure cases of our model in which the targets are not detected.
We observed that similar cases also occur in the EchoFusion\cite{echofusion} and DPFT\cite{dpft}. 
A common characteristic of these failures is that the vehicles are oriented perpendicular to the ego vehicle’s driving direction, 
while most training samples feature vehicles parallel to the ego motion along the road. 
This data imbalance in the training set is the primary cause of such missed detections.
We will explore more effective data augmentation and training strategies to address this issue.
Moreover, as mentioned in \cite{dpft}, the K-Radar dataset, particularly revision v1.0, 
suffers from sensor calibration issues due to intrinsic and extrinsic parameter misalignment. 
As shown in the left example of Fig.~\ref{vis_failure}, the ground-truth bounding box projection on the image is noticeably offset, 
which further contributes to performance degradation of these detection models.

\section{Conclusion and Future Work}
This paper proposes RadarXFormer, a 3D object detection framework that fuses 4D radar spectra with RGB images to achieve robust perception under adverse weather conditions. 
By constructing an efficient 3D radar representation and introducing a cross-dimension (3D-2D) fusion mechanism based on multi-scale deformable attention, 
RadarXFormer effectively reduces data complexity while preserving critical spatial and Doppler information. 
Experiments on the K-Radar dataset demonstrate that the proposed framework achieves improved detection accuracy and robustness, 
significantly outperforming existing radar-camera fusion methods. 
These results confirm the strong potential of 4D radar spectra and camera fusion for all-weather autonomous driving. 

Despite the great potential of 4D radar spectral data for 3D object detection, this work also raises several challenges. 
Although we propose a denoising method to suppress noise and compress the radar spectrum, its performance remains limited. 
Therefore, future work will focus on developing more efficient denoising and data-compression strategies for raw radar spectra that 
better balance the preservation of critical information with data size and computational complexity, 
particularly for radar spectrum representations. 
In addition, considering the data distribution imbalance in existing radar datasets, 
future work should place greater emphasis on increasing data diversity and developing effective data augmentation techniques for radar data.



\bibliographystyle{IEEETran}
\bibliography{references.bib} 

\end{document}